%% file: main.tex
\documentclass[10pt,twocolumn,letterpaper]{article}

\usepackage{cvpr}              

\usepackage{graphicx}
\usepackage{amsmath}
\usepackage{amssymb}
\usepackage{booktabs}
\usepackage{multirow}
\usepackage{colortbl}
\definecolor{Gray}{gray}{0.95}
\makeatletter
\newcommand{\thickhline}{%
 \noalign {\ifnum 0=`}\fi \hrule height 1pt
 \futurelet \reserved@a \@xhline
}
\makeatother

\usepackage[pagebackref,breaklinks,colorlinks]{hyperref}

\usepackage[capitalize]{cleveref}
\crefname{section}{Sec.}{Secs.}
\Crefname{section}{Section}{Sections}
\Crefname{table}{Table}{Tables}
\crefname{table}{Tab.}{Tabs.}

\def\cvprPaperID{5880} 
\def\confName{CVPR}
\def\confYear{2022}

\makeatletter
\newcommand{\printfnsymbol}[1]{%
  \textsuperscript{\@fnsymbol{#1}}%
}
\makeatother

\begin{document}

\title{3D-SPS: Single-Stage 3D Visual Grounding via \\ Referred Point Progressive Selection}

\author{Junyu Luo$^{1,2}$~$^{*}$, Jiahui Fu$^{1,2}$~\thanks{Equal contribution}~, Xianghao Kong$^{1,2}$~, Chen Gao$^{1,2}$~\thanks{Corresponding author: \textit{Chen Gao~(gaochen.ai@gmail.com)}.}~, \\
Haibing Ren$^{3}$~, Hao Shen$^{3}$~, Huaxia Xia$^{3}$~, Si Liu$^{1,2}$~\\
\small{$^1$Institute of Artificial Intelligence, Beihang University} \ 
\small{$^2$Hangzhou Innovation Institute, Beihang University} \ 
\small{$^3$Meituan Inc.}\\
}

\maketitle

\input{./tex/0_abstract.tex}

\input{./tex/1_intro.tex}

\input{./tex/2_related.tex}

\input{./tex/3_method.tex}


\input{./tex/4_exp.tex}


\input{./tex/5_conclusion.tex}

{\small
\bibliographystyle{ieee_fullname}
\bibliography{egbib}
}

\end{document}

%% file: tex/0_abstract.tex
\begin{abstract}
3D visual grounding aims to locate the referred target object in 3D point cloud scenes according to a free-form language description.
Previous methods mostly follow a two-stage paradigm, i.e., language-irrelevant detection and cross-modal matching, which is limited by the isolated {architecture}.
In such a paradigm, the detector needs to sample keypoints from raw point clouds due to the inherent properties of 3D point clouds (irregular and large-scale), to generate the corresponding object proposal for each keypoint.
However, sparse proposals may leave out the target in detection, while dense proposals may confuse the matching model.
Moreover, the language-irrelevant detection stage can only sample a small proportion of keypoints on the target, deteriorating the target prediction.
In this paper, we propose a \textbf{3D} \textbf{S}ingle-Stage Referred Point \textbf{P}rogressive \textbf{S}election (3D-SPS) method, which progressively selects keypoints with the guidance of language and directly locates the target.
Specifically, we propose a Description-aware Keypoint Sampling~(DKS) module to coarsely focus on the points of language-relevant objects, which are significant clues for grounding. 
Besides, we devise a Target-oriented Progressive Mining~(TPM) module to finely concentrate on the points of the target, which is enabled by progressive intra-modal relation modeling and inter-modal target mining.
3D-SPS bridges the gap between detection and matching in the 3D visual grounding task, localizing the target at a single stage.
Experiments demonstrate that 3D-SPS achieves state-of-the-art performance on both ScanRefer and Nr3D/Sr3D datasets.
\end{abstract}

%% file: tex/1_intro.tex
\vspace{-2mm}
\section{Introduction}
\label{sec:intro}
\begin{figure}[t]
    \centering 
    \includegraphics[width=1\linewidth]{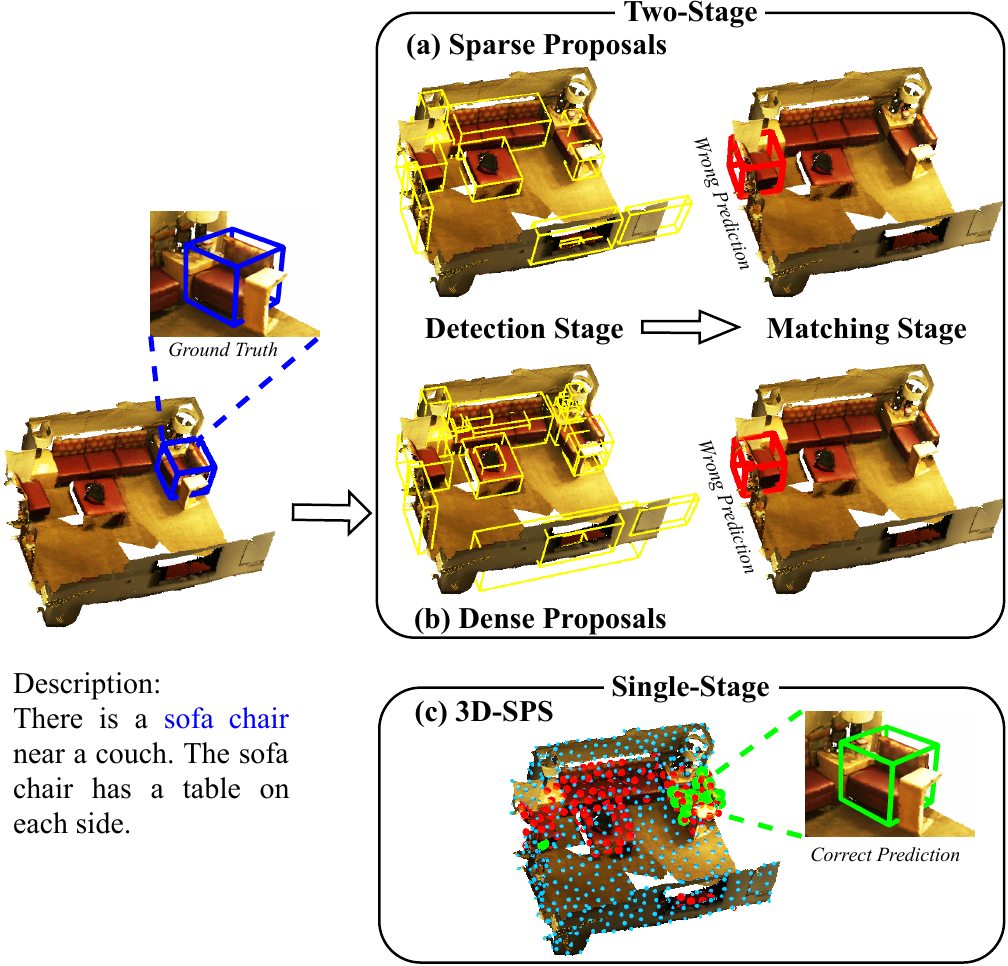}
    \caption{Traditional two-stage 3D VG methods are limited by the isolation of the detection stage and the matching stage. (a)~Sparse \mbox{proposals} may leave out the target in detection. (b)~Dense \mbox{proposals} could confuse the matching model. (c)~3D-SPS progressively selects keypoints~(\textcolor{blue}{blue} points$\rightarrow$\textcolor{red}{red} points$\rightarrow$\textcolor{green}{green} points) and performs referring at a single stage. Noted that dense surfaces are utilized only to help readers understand the example 3D scene, while the input of our method only contains sparse point clouds.}
    \label{fig:figure1}\vspace{-2mm}
\end{figure}
Visual Grounding~(VG) aims to localize the target object in the scene based on an object-related linguistic description. 
In recent years, the 3D VG task has received increasing attention owing to its wide applications, such as autonomous robots and human-machine interaction in AR/VR/Metaverse.
Even though much progress~\cite{yang2019dynamic, yang2019cross, wang2019neighbourhood, wang2018learning, zhang2018grounding, yu2018mattnet, yu2016modeling,sadhu2019zero, yang2019fast, yang2020improving} has been achieved in the 2D VG task, it is still challenging to locate the referred target object in 3D scenes since point clouds are irregular and large-scale.

Existing 3D VG methods~\cite{scanrefer, Yuan_2021_ICCV, sat, Zhao_2021_ICCV, he2021transrefer3d, Feng_2021_ICCV} are mainly based on the \emph{detection-then-matching} two-stage pipeline.
The first stage is language-irrelevant detection, where general 3D object detectors~\cite{DBLP:conf/iccv/QiLHG19, groupfree, DBLP:conf/cvpr/ChengSSY021} are adopted to produce numerous object proposals. The second stage is cross-modal matching, where specific vision-language attention mechanisms are usually designed to match the proposal and  the description.
Previous methods primarily focus on the second stage, \ie, exploring relations among proposals to distinguish the target object.

We argue that the separation of the two stages limits the existing methods. 
Previous 2D detection methods adopt data-independent anchor boxes as proposals on regular and well-organized images. 
However, the anchor-based fashion is generally impractical for the large-scale and irregular 3D point clouds. 
Consequently, the 3D detector utilized in the first stage needs to sample a limited number of keypoints to represent the whole scene and generate the corresponding proposal for each keypoint. 
However, sparse proposals may leave out the target in the detection stage (\eg, the \emph{sofa chair} in Figure~\ref{fig:figure1}~(a)), which leads to the inability to locate the target in the matching stage.
Meanwhile, dense proposals may contain redundant objects, causing the inter-proposal relationship so complex that the matching module struggles to distinguish the target.
As shown in Figure~\ref{fig:figure1}~(b), it is difficult to select the right \emph{sofa chair} from these numerous proposals with similar appearances.
Therefore, the two-stage grounding methods face a dilemma of deciding the proposal number.
Besides, the keypoint sampling strategy~(\eg, Farthest Point Sampling~(FPS)~\cite{pointnet++}) usually adopted in the detector at the first stage is also language-irrelevant. The strategy aims to sample keypoints to cover the entire scene as much as possible to detect all potential objects. Thus, the proportion of target keypoints is relatively small, which is unfavorable for the target prediction.

To address the aforementioned issues, we propose a \textbf{3D} \textbf{S}ingle-Stage Referred Point \textbf{P}rogressive \textbf{S}election~(3D-SPS) method in this paper. 
Our main idea is to progressively select keypoints under the guidance of the language description throughout the whole process, as shown in Figure~\ref{fig:figure1}~(c).
Based on this idea, we propose a Description-aware Keypoint Sampling~(DKS) module to coarsely focus on the points of language-relevant objects, \eg, \emph{sofa chair, couch, and table} in Figure~\ref{fig:figure1}~(c).
These keypoints provide significant clues for localizing the grounding target in the following cross-modal interaction. 
Besides, we devise a Target-oriented Progressive Mining~(TPM) module, which conducts progressive mining to finely figure out the target.
We leverage the self/cross-attention mechanism to model intra/inter-modal relationships respectively.
In addition, we fuse the keypoint features with point features of the whole scene to achieve global localization perception.
To progressively select keypoints of the target, we utilize the language-points cross-attention map to select the keypoints that the language pays more attention to and discard irrelevant points.
The model gradually concentrates on the target and obtains a condensed set of keypoints through multiple layers.
Thus, the proportion of target points will gradually increase with richer target-related features, which benefits the target box regression.
Finally, 3D-SPS distinguishes the target from the condensed keypoint set and regresses its bounding box.
Note that 3D-SPS is also consistent with the commonsense of how human finds the target object. Commonly, a human first selects a coarse candidate set according to the language description and then finely recognize and judge it to select the target object.~\cite{jacob2021qualitative, ullman2016atoms}

In summary, we make the following contributions:
\vspace{-4pt}
\begin{itemize} 
\setlength\itemsep{-1pt}
\item We propose the 3D-SPS method, which directly performs 3D VG at a single stage to bridge the gap between detection and matching.
To the best of our knowledge, 3D-SPS is the first work investigating single-stage 3D VG.

\item We treat the 3D VG task as a keypoint selection problem. Two selection modules, \ie, DKS and TPM, are designed to progressively select target-related keypoints. DKS samples the coarse language-relevant keypoints, and TPM finely mines the cross-modal relationship to distinguish the target.

\item Extensive experiments confirm the effectiveness of our method. 3D-SPS achieves the state-of-the-art performance on both \textit{ScanRefer}~\cite{scanrefer} and \textit{Nr3D/Sr3D}~\cite{achlioptas2020referit3d} datasets. The code is provided in \url{https://github.com/fjhzhixi/3D-SPS}.

\vspace{-3pt}
\end{itemize}

%% file: tex/2_related.tex
\begin{figure*}[ht]
    \centering 
    \includegraphics[width=0.9\textwidth]{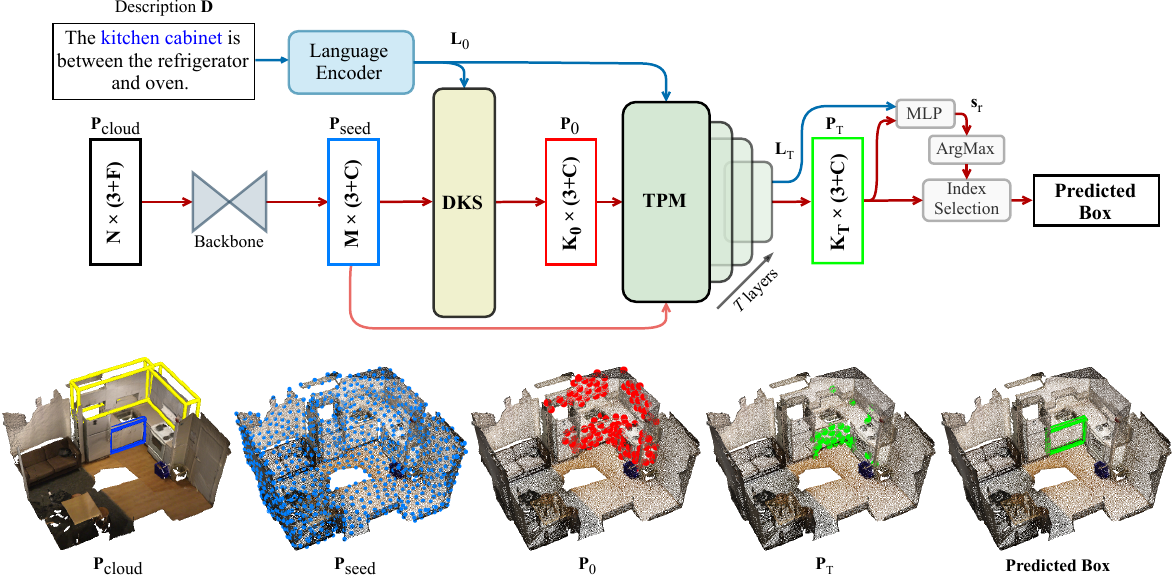}\vspace{-2.5mm}
    \caption{\textbf{3D-SPS framework.} 
    We take the 3D VG task as a keypoint selection problem and avoid the separation of detection and matching.
    Specifically, we use PointNet++ as the backbone to extract point seeds $\mathbf{P}_{seed}$ from  $\mathbf{P}_{cloud}$.
    After that, we coarsely sample the language-relevant keypoints $\mathbf{P}_{0}$ by DKS with word features $\mathbf{L}_0$, which are mostly on the \textit{kitchen cabinets}, \textit{refrigerator} and \textit{oven} in the figure.
    Then, TPM finely selects target keypoints $\mathbf{P}_{T}$ and predict referring confidence scores $\mathbf{s}_r$. Here the keypoints are concentrated on the target \textit{kitchen cabinet}.
    Finally, the target box is regressed from the keypoint with the highest $\mathbf{s}_r$ in $\mathbf{P}_{T}$. 
    The \textcolor{blue}{blue} box is the ground truth.
    The \textcolor{yellow}{yellow} boxes are objects of the same category as the target.
    The \textcolor{green}{green} box is our target prediction.
    Best viewed in color.}\vspace{-2.5mm}
    \label{fig:framework}
\end{figure*}

\section{Related Work}
\vspace{0.5mm}
\noindent\textbf{Visual Grounding on 2D Images.}
The goal of visual grounding on 2D images is to select a referred target according to the referring expression~\cite{hu2016natural, yu2018mattnet, nagaraja2016modeling, gao2021room}.
Two mainstream frameworks have been proposed in succession: two-stage and one-stage methods. Specifically, two-stage methods~\cite{yang2019dynamic, yang2019cross, wang2019neighbourhood, wang2018learning, zhang2018grounding, yu2018mattnet, yu2016modeling, hong2019learning, liu2019learning, zhuang2018parallel} first generate region proposals with object detectors and then select the target region by matching the language features with the proposals.
Each proposal is treated the same in the matching stage, despite their importance in the referring context varies.
Besides, one-stage methods~\cite{sadhu2019zero, yang2019fast, yang2020improving, chen2018real, deng2021transvg, liao2020real} eliminate the proposal generation and feature extraction stage in two-stage frameworks.
In these methods, linguistic features are densely fused with each pixel or patch to generate multi-modal feature maps for regressing the bounding box.

However, one-stage methods in 2D VG could not be directly lifted to 3D VG. Firstly, 3D point clouds are numerous and noisy.
Therefore, it is computationally unacceptable~\cite{zhou2018voxelnet, sparseconv, graham2017submanifold} to treat each point as a candidate.
Then, due to the large-scale and complexity of 3D scenes, it is not easy to model the relationship of all objects and figure out the target\cite{Zhao_2021_ICCV, he2021transrefer3d, sat}.
Moreover, 2D one-stage methods adopt the sliding-window manner like~\cite{he2016deep, simonyan2014very}, which cannot deal with 3D points since 2D input is highly regular while 3D points are inherently sparse, unordered, and irregular~\cite{qi2017pointnet, pointnet++}.
In this paper, we propose 3D-SPS to address the problems introduced by 3D point clouds, which becomes the leading 3D VG solution.
    
\vspace{0.5mm}
\noindent\textbf{Visual Grounding on 3D Point Clouds.}
With the prevalence of deep learning technologies on 3D point clouds, the 3D VG task has attracted much attention.
Chen \textit{et al.}~\cite{scanrefer} released a 3D VG dataset \textit{ScanRefer}, in which the bounding boxes of objects are referred by their corresponding descriptions in an indoor scene.
ReferIt3D~\cite{achlioptas2020referit3d} also proposes two datasets, \ie, \textit{Sr3D} and \textit{Nr3D}, for the 3D VG task.

Existing 3D VG works~\cite{scanrefer, DBLP:conf/aaai/HuangLCL21, Yuan_2021_ICCV, Zhao_2021_ICCV, sat, he2021transrefer3d,Feng_2021_ICCV, roh2021languagerefer} mainly focus on better modeling the relationship among objects to locate the target object, \eg, adopting graph neural network~\cite{DBLP:conf/aaai/HuangLCL21}, and attention mechanisms~\cite{Zhao_2021_ICCV}.
To the best of our knowledge, previous 3D grounding approaches can generally be concluded into a detection-then-matching two-stage framework.
In these methods, the detection stage fails to leverage the language context to concentrate on the points that are more essential to the referring task.
To overcome those shortcomings, we propose the first single-stage method in 3D VG to progressively select keypoints under the guidance of the description.

%% file: tex/3_method.tex
\section{Method}
In this section, we detail the 3D-SPS method.
In Sec~\ref{sec:overview}, we present an overview of 3D VG task and our method.
In Sec~\ref{sec:method-2} and Sec~\ref{sec:method-3}, we dive into the technical details and how we obtain the target by progressive keypoint selection.
In Sec~\ref{sec:method-4}, we introduce the training objectives of 3D-SPS.

\subsection{Overview}
\label{sec:overview} 
In the 3D VG task, the inputs are the point clouds $\mathbf{P}_{cloud}\in \mathbb{R}^{N\times(3+F)}$ and a free-form  plain text description $\mathbf{D}$ of the target object with $W$ words, where $\mathbf{P}_{cloud}$ contains 3D coordinates and $F$-dimensional auxiliary feature~(RGB, normal vectors, \etc.) of $N$ points.
The goal of this task is to locate the target object~(\ie, the most relevant object to the description) and predict its bounding box.

The main idea of 3D-SPS is the progressive keypoint selection process, as shown in Figure~\ref{fig:framework}.
\textit{Firstly}, we adopt a widely used PointNet++~\cite{pointnet++} as the backbone network to extract point features from $\mathbf{P}_{cloud}$. 
The backbone outputs $M$ seed points with $(x, y, z)$ coordinates and $C$-dimensional enriched local features $\mathbf{P}_{seed} \in \mathbb{R}^{M \times (3+C)}$.
Meanwhile, we use the language encoder to extract $H$-dimensional word features $\mathbf{L}_0 \in \mathbb{R}^{W \times H}$ from $W$-length description $\mathbf{D}$.
\textit{Secondly}, DKS module selects $K_{0}$ language-relevant keypoints with features $\mathbf{P}_{0} \in \mathbb{R}^{K_0 \times (3+C)}$ from $M$ seed points based on word features $\mathbf{L}_0$.
These keypoints belong to the objects whose categories are mentioned in the description, providing significant clues to distinguishing the grounding target.
\textit{Thirdly}, TPM module takes point features $\mathbf{P}_{0}$ and word features $\mathbf{L}_0$ as inputs. The $t$-th layer of the TPM module takes $\mathbf{P}_{t-1}$ and $\mathbf{L}_{t-1}$ as inputs and outputs $\mathbf{P}_{t}$ and $\mathbf{L}_{t}$. TPM progressively distinguishes the grounding target by multi-layer cross-modal transformers. We select $K_T$ keypoints with features $\mathbf{P}_{T} \in \mathbb{R}^{K_T \times (3+C)}$ and update the word features as $\mathbf{L}_T$.
\textit{Lastly}, we predict the referring confidence score $\mathbf{s}_r$ based on keypoint features $\mathbf{P}_{T}$ and cross-modally aligned word features $\mathbf{L}_{T}$ by a simple MLP head. The keypoint feature with the highest $\mathbf{s}_r$ is used to regress the bounding box of the grounding target as the center $\mathbf{c} \in \mathbb{R}^3$ and the size $\mathbf{s}\in \mathbb{R}^3$.
By treating the 3D VG task as a keypoint selection problem, our 3D-SPS concentrates on distinguishing the keypoints of the target object from point clouds for predicting the bounding box directly, which is more effective than traditional detection-then-matching two-stage methods.

\begin{figure}[t]
    \centering 
    \includegraphics[width=0.45\textwidth]{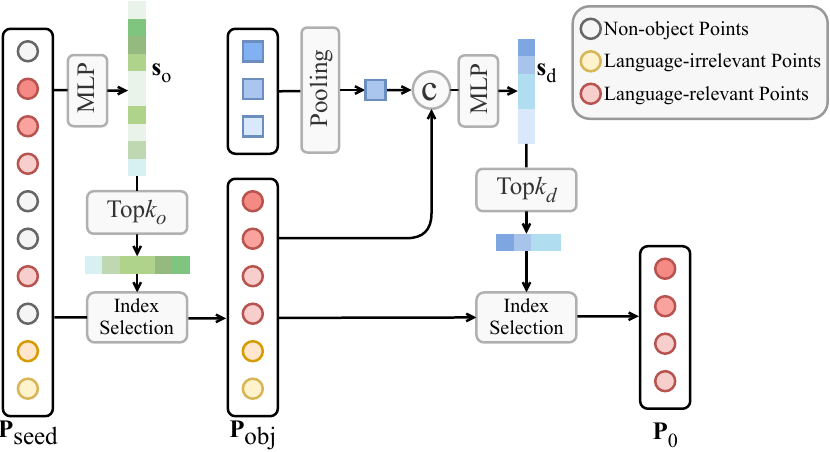}\vspace{-2.5mm}
    \caption{\textbf{The DKS module.} We use object confidence score $\mathbf{s}_{o}$ to select points near object centers and description relevance score $\mathbf{s}_{d}$ to select language-relevant points.}
    \label{fig:method-DKS}\vspace{-2.5mm}
\end{figure}

\subsection{Description-aware Keypoint Sampling}

Since the search space of 3D anchor boxes is huge, the data-independent anchor assignment strategy  widely adopted in 2D object detection~\cite{DBLP:conf/nips/RenHGS15} is impractical when lifted to 3D~\cite{groupfree}.
\label{sec:method-2}
\begin{figure}[t]
    \centering 
    \includegraphics[width=0.4\textwidth]{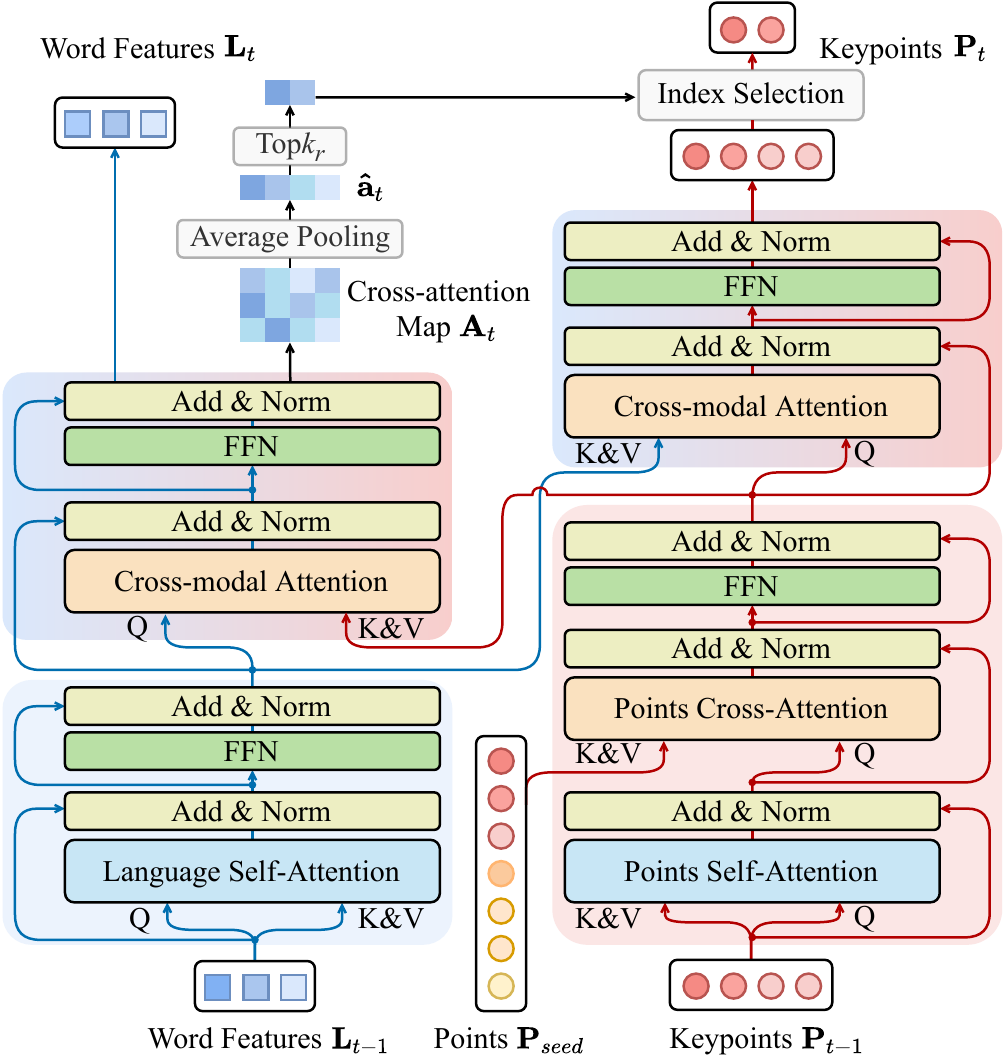}\vspace{-2.5mm}
    \caption{\textbf{The TPM module.} It is a two-stream cross-modal transformer model. We select the keypoints of the target based on the language-points cross-attention map $\mathbf{A}_t$ at the $t$-th layer.}
    \label{fig:method-RKS}\vspace{-2.5mm}
\end{figure}
To this end, most 3D object detection methods~\cite{DBLP:conf/iccv/QiLHG19, groupfree, DBLP:conf/cvpr/ChengSSY021} usually adopt sampling methods~(\eg, FPS~\cite{pointnet++}) to sample keypoints from seed points and generate a proposal for each selected point.
Existing detection-then-matching methods for the 3D VG task usually use the same strategy at the detection stage.
However, directly adopting the sampling strategy in detection to the 3D VG task is not sensible because of the divergence of interest of the two tasks.
The sampling objective of 3D object detection is to cover the entire scene as much as possible for detecting potential objects, while the goal of 3D VG is to locate the referred target.
Therefore, we propose DKS to help the model focus on the keypoints of language-relevant objects instead of the whole scene.
Specifically, we bring word features into the sampling process to select keypoints of the objects whose categories are mentioned in the description.
These keypoints contain the information of not only the target object but also related objects to help determine the target.

Figure~\ref{fig:method-DKS} details the DKS.
We first obtain an object confidence score $\mathbf{s}_o$ based on point features $\mathbf{P}_{seed}$ to clarify whether the point is near an object center. 
The keypoint features $\mathbf{P}_{obj}$ with top $k_o$ highest $\mathbf{s}_o$ are selected as:
\begin{equation}
\begin{aligned}
    \mathbf{s}_o &= \mathtt{MLP}(\mathbf{P}_{seed}), \\
    \mathbf{P}_{obj} &= \mathbf{P}_{seed}\left[ \mathtt{argtopk}(\mathbf{s}_o, k_o) \right].
\end{aligned}
\end{equation}

Then a description relevance score $\mathbf{s}_d$ is utilized to select top $k_d$ keypoints as $\mathbf{P}_{0}$ that are related to the description context $\mathbf{L}_0$.
We jointly use point features $\mathbf{P}_{obj}$ and global word features to predict the $\mathbf{s}_d$  of each point, which can be formulated as:
\begin{equation}
\begin{aligned}
    \mathbf{s}_d &= \mathtt{MLP}(\mathbf{P}_{obj} \  || \  \mathtt{MaxPool}(\mathbf{L}_0)), \\
    \mathbf{P}_{0} &= \mathbf{P}_{obj}\left[ \mathtt{argtopk}(\mathbf{s}_d, k_d) \right].
\end{aligned}
\end{equation}

\subsection{Target-oriented Progressive Mining}
\label{sec:method-3}

With the coarsely selected language-relevant keypoints by DKS, we perform fine target mining with the TPM module.
TPM is constructed by a $T$-layer stacked multi-modal two-stream transformer model, where both word features and keypoint features are processed in separate streams and interact through cross-modal attention layers to model the relationship and mine the target.
At the $t$-th layer, TPM selects $\mathbf{P}_t$ from $\mathbf{P}_{t-1}$. 
TPM progressively selects the keypoints and concentrates the attention by discarding target-irrelevant keypoints in each layer.
\vspace{0.5mm}
\noindent\textbf{Intra/inter-modal Modeling.}
As Figure~\ref{fig:method-RKS} shows, we employ the attention mechanism~\cite{vaswani2017attention} to learn intra-modal relationships.
For point features, the point self-attention block helps to refine point visual features and exploits their spatial relationship. 
For word features, the language self-attention block is used to extract context relationships.
Specially, we leverage a point cross-attention block to model the global location of keypoints in the scene because the interaction of selected keypoints could not well model descriptions which include the global location like ``\textit{in the center/corner of room}".
Therefore, the scene point clouds $\mathbf{P}_{seed}$~(point features before DKS) are fused to acquire global scene features.
Next, point features and word features interact in cross-modal attention blocks. 
In these blocks, the points branch is assisted by word features to distinguish the target, while the language branch fuses the scene information by attending to point features.

\vspace{0.5mm}
\noindent\textbf{Attention-guided Keypoint Selection.}
TPM reduces the keypoint set at each layer and gradually focuses on the target, as shown in Figure~\ref{fig:method-RKS}. 
We make use of the language-points cross-attention map $\mathbf{A}_{t}$, which represents the importance of keypoints to the referring task.
Specifically, we perform average pooling on $\mathbf{A}_{t}$ and obtain point-wise attention scores $\mathbf{\hat{a}}_t \in \mathbb{R}^{K_{t-1}}$.
Then the keypoints with top $k_r$ highest $\mathbf{\hat{a}}_t$ are selected for the next layer as follow:
\begin{equation}
\begin{aligned}
    \mathbf{\hat{a}}_t &= \mathtt{AvgPool}(\mathbf{A}_t), \\
    \mathbf{P}_{t} &= \mathbf{P}_{t-1}\left[ \mathtt{argtopk}(\mathbf{\hat{a}}_t, k_r) \right].
\end{aligned}
\end{equation}

\subsection{Training Objectives}
\label{sec:method-4}
\vspace{0.5mm}
\noindent\textbf{Visual Grounding Loss.}~3D VG loss $\mathcal{L}_{\mathrm{VG}}$ is the primary loss of our framework.
In the training phase, we supervise referring confidence scores $\mathbf{s}_r$ predicted from $\mathbf{P}_T$ with the target label.
During inference, we only choose the keypoint with the highest $\mathbf{s}_r$ from $\mathbf{P}_T$ to predict the target box.
We adapt the loss in ScanRefer~\cite{scanrefer} to our framework.
In ScanRefer, the target label of $\mathbf{s}_r$ is a one-hot label.
The keypoint whose proposal box has the highest IoU with the ground truth target box is set to $1$, and others are set to $0$.
However, in 3D-SPS, we usually obtain several feasible keypoints of the target after TPM since the model aims to select points on it.
Therefore, we modify this target label from one-hot to multi-hot.
Specifically, we assign $1$ to keypoints whose predicted boxes' IoUs with the ground truth target box are the top $k_1$ highest and greater than the threshold $\theta$.

\vspace{0.5mm}
\noindent\textbf{DKS Loss.}~In the DKS module, we apply $\mathcal{L}_{\mathrm{DKS}}$ to supervise the object confidence score $\mathbf{s}_o$ and the description relevance score $\mathbf{s}_d$ with Focal Loss~\cite{Lin_2017_ICCV}.
The $\mathbf{s}_o$ is supervised by whether the point is inside an object box and belongs to the $k_2$-closest points to the object center.
The $\mathbf{s}_d$ is supervised by whether the point belongs to any object whose category is mentioned in the description.

\vspace{0.5mm}
\noindent\textbf{Detection Loss.}~Following the loss used in \cite{DBLP:conf/iccv/QiLHG19, groupfree}, we use the object detection loss $\mathcal{L}_{\mathrm{Det}}$ as an auxiliary loss for VG task.
Specifically, $\mathcal{L}_{\mathrm{Det}}$ comprises object semantic classification loss $\mathcal{L}_{\mathrm{Cls}}$, objectness binary classification loss $\mathcal{L}_{\mathrm{Obj}}$, center offset regression loss $\mathcal{L}_{\mathrm{Center}}$, and bounding box regression loss $\mathcal{L}_{\mathrm{Box}}$.
In the training phase, we supervise the box of objects predicted by all keypoints of each TPM layer.
During inference, we only use the box prediction of the keypoint with the highest $\mathbf{s}_r$ from the last TPM layer as our predicted grounding target.

\vspace{0.5mm}
\noindent\textbf{Language Classification Loss.}~Following ~\cite{scanrefer}, we also introduce the language classification loss $\mathcal{L}_{\mathrm{Lang}}$ as an auxiliary loss, which includes a multi-class object classification loss for the target category based on the updated language features of each TPM layer.

In summary, the total loss is:
$\mathcal{L} = \alpha_1 \mathcal{L}_{\mathrm{VG}} + \alpha_2 \mathcal{L}_{\mathrm{DKS}} + \alpha_3 \mathcal{L}_{\mathrm{Det}} + \alpha_4 \mathcal{L}_{\mathrm{Lang}}$,
where the weights $\alpha_1, \alpha_2, \alpha_3, \alpha_4$ are used for balancing different loss terms.

%% file: tex/4_exp.tex
\section{Experiments}

\begin{table*}[t]
	\centering
	\vspace{.1cm}
	\small
	\begin{tabular}{lcc|cc|cc|cc}
\toprule
\multirow{2}*{Method} & \multirow{2}*{Pub.} & \multirow{2}*{Input} & \multicolumn{2}{c|}{Unique} & \multicolumn{2}{c|}{Multiple} & \multicolumn{2}{c}{Overall}\\
~ & ~ & ~ & Acc@0.25 & Acc@0.5 & Acc@0.25 & Acc@0.5 & Acc@0.25 & Acc@0.5 \\
\hline  
\hline
SCRC~\cite{hu2016natural} & CVPR16 & 2D only & 24.03 & 9.22 & 17.77 & 5.97 & 18.70 & \cellcolor{Gray}6.45 \\
One-stage~\cite{yang2019fast}& ICCV19 & 2D only & 29.32 & 22.82 & 18.72 & 6.49 & 20.38 & \cellcolor{Gray}9.04 \\
\hline
ScanRefer~\cite{scanrefer} & ECCV20 & 3D only & 67.64 & 46.19 & 32.06 & 21.26 & 38.97 & \cellcolor{Gray}26.10 \\
TGNN~\cite{DBLP:conf/aaai/HuangLCL21} & AAAI21 & 3D only & 68.61 & 56.80 & 29.84 & 23.18 & 37.37 & \cellcolor{Gray}29.70 \\
IntanceRefer~\cite{Yuan_2021_ICCV} & ICCV21 & 3D only & {77.45} & \textbf{66.83} & 31.27 & 24.77 & 40.23 & \cellcolor{Gray}32.93 \\
SAT~\cite{sat} & ICCV21 & 3D only & 73.21 & 50.83 & 37.64 & 25.16 & 44.54 & \cellcolor{Gray}30.14 \\
3DVG-Transformer~\cite{Zhao_2021_ICCV} & ICCV21 & 3D only & 77.16 & 58.47 & {38.38} & {28.70} & {45.90} & \cellcolor{Gray}{34.47} \\
\textbf{3D-SPS~(Ours)} & - & 3D only & \textbf{81.63} & {64.77} & \textbf{39.48} & \textbf{29.61} & \textbf{47.65} & \cellcolor{Gray}\textbf{36.43} \\
\hline
ScanRefer~\cite{scanrefer} & ECCV20 & 2D + 3D & 76.33 & 53.51 & 32.73 & 21.11 & 41.19 & \cellcolor{Gray}27.40 \\
InstanceRefer~\cite{Yuan_2021_ICCV} & ICCV21 & 2D + 3D & 75.72 & {64.66} & 29.41 & 22.99 & 38.40 & \cellcolor{Gray}31.08 \\
3DVG-Transformer~\cite{Zhao_2021_ICCV} & ICCV21 & 2D + 3D & {81.93} & 60.64 & {39.30} & {28.42} & {47.57} & \cellcolor{Gray}{34.67} \\
\textbf{3D-SPS~(Ours)} & - & 2D + 3D & \textbf{84.12} & \textbf{66.72} & \textbf{40.32} & \textbf{29.82} & \textbf{48.82} & \cellcolor{Gray}\textbf{36.98} \\
	\bottomrule 
	\end{tabular}\vspace{-1mm}
	\caption{\textbf{Comparison on \textit{ScanRefer}.} The \textit{unique} stands for samples with no distracting objects and \textit{multiple} for remaining samples. We measure the percentage of predictions whose IoU with the ground truth is greater than $\{0.25, 0.5\}$. }
	\label{tab:scanref-all-methods}
\end{table*}

\begin{table*}[ht]
	\centering 
	\footnotesize
    \vspace{.1cm}
	\small
	\begin{tabular}{lc|c|c|c|c|c}
	\toprule  
Method & Pub. & Easy & Hard & View-dep. & View-indep. & Overall \\
\hline
\hline
\multicolumn{7}{c}{\textbf{Nr3D}} \\
\hline
ReferIt3DNet~\cite{achlioptas2020referit3d} & ECCV20 & {43.6\% $\pm$ 0.8\%} & {27.9\% $\pm$ 0.7\%} & {32.5\% $\pm$ 0.7\%} & {37.1\% $\pm$ 0.8\%} & \cellcolor{Gray}{35.6\% $\pm$ 0.7\%} \\
TGNN~\cite{DBLP:conf/aaai/HuangLCL21} & AAAI21 & {44.2\% $\pm$ 0.4\%} & {30.6\% $\pm$ 0.2\%} & {35.8\% $\pm$ 0.2\%} & {38.0\% $\pm$ 0.3\%} & \cellcolor{Gray}{37.3\% $\pm$ 0.3\%} \\
IntanceRefer~\cite{Yuan_2021_ICCV} & ICCV21 & {46.0\% $\pm$ 0.5\%} & {31.8\% $\pm$ 0.4\%} & {34.5\% $\pm$ 0.6\%} & {41.9\% $\pm$ 0.4\%} & \cellcolor{Gray}{38.8\% $\pm$ 0.4\%} \\
3DVG-Transformer~\cite{Zhao_2021_ICCV} & ICCV21 & {48.5\% $\pm$ 0.2\%} & {34.8\% $\pm$ 0.4\%} & {34.8\% $\pm$ 0.7\%} & {43.7\% $\pm$ 0.5\%} & \cellcolor{Gray}{40.8\% $\pm$ 0.2\%}\\
LanguageRefer~\cite{roh2021languagerefer} & CoRL21 & {51.0\%} & {36.6\%} & {41.7\%} & {45.0\%} & \cellcolor{Gray}{43.9\%}\\
SAT~\cite{sat} & ICCV21 & {56.3\% $\pm$ 0.5\%} & {42.4\% $\pm$ 0.4\%} & {46.9\% $\pm$ 0.3\%} & {50.4\% $\pm$ 0.3\%} & \cellcolor{Gray}{49.2\% $\pm$ 0.3\%}\\
\textbf{3D-SPS~(Ours)} & - & \textbf{{58.1\% $\pm$ 0.3\%}} & \textbf{{45.1\% $\pm$ 0.4\%}} & \textbf{{48.0\% $\pm$ 0.2\%}} & \textbf{{53.2\% $\pm$ 0.3\%}} & \cellcolor{Gray}\textbf{{51.5\% $\pm$ 0.2\%}} \\

\hline
\multicolumn{7}{c}{\textbf{Sr3D}} \\
\hline

ReferIt3DNet~\cite{achlioptas2020referit3d} & ECCV20 & {44.7\% $\pm$ 0.1\%} & {31.5\% $\pm$ 0.4\%} & {39.2\% $\pm$ 1.0\%} & {40.8\% $\pm$ 0.1\%} & \cellcolor{Gray}{40.8\% $\pm$ 0.2\%} \\
TGNN~\cite{DBLP:conf/aaai/HuangLCL21} & AAAI21 & {48.5\% $\pm$ 0.2\%} & {36.9\% $\pm$ 0.5\%} & {45.8\% $\pm$ 1.1\%} & {45.0\% $\pm$ 0.2\%} & \cellcolor{Gray}{45.0\% $\pm$ 0.2\%} \\
IntanceRefer~\cite{Yuan_2021_ICCV} & ICCV21 & {51.1\% $\pm$ 0.2\%} & {40.5\% $\pm$ 0.3\%} & {45.4\% $\pm$ 0.9\%} & {48.1\% $\pm$ 0.3\%} & \cellcolor{Gray}{48.0\% $\pm$ 0.3\%} \\
3DVG-Transformer~\cite{Zhao_2021_ICCV} & ICCV21 & {54.2\% $\pm$ 0.1\%} & {44.9\% $\pm$ 0.5\%} & {44.6\% $\pm$ 0.3\%} & {51.7\% $\pm$ 0.1\%} & \cellcolor{Gray}{51.4\% $\pm$ 0.1\%} \\
LanguageRefer~\cite{roh2021languagerefer} & CoRL21 & \textbf{{58.9\%}} & {49.3\%} & {49.2\%} & {56.3\%} & \cellcolor{Gray}{56.0\%}\\
SAT~\cite{sat} & ICCV21 & {-} & {-} & {-} & {-} & \cellcolor{Gray}{57.9\% $\pm$ 0.1\%}\\
\textbf{3D-SPS~(Ours)} & - & {56.2\% $\pm$ 0.6\%} & \textbf{{65.4\% $\pm$ 0.1\%}} & \textbf{49.2\% $\pm$ 0.5\%} & \textbf{{63.2\% $\pm$ 0.2\%}} & \cellcolor{Gray}\textbf{{62.6\% $\pm$ 0.2\%}} \\
\bottomrule 
	\end{tabular}\vspace{-1mm}
	\caption{\textbf{Comparison on \textit{Nr3D} and \textit{Sr3D}.} \textit{Easy} samples contain no distractor, and the remaining belong to \textit{Hard}. \textit{View-dep.}/\textit{View-indep.} refer to whether the description is dependent or independent on the camera view.}
	\label{tab:referit3d-all-methods}\vspace{-2mm}
\end{table*}

\subsection{Datasets}
\vspace{0.5mm}
\noindent\textbf{ScanRefer.}~The \textit{ScanRefer} dataset~\cite{scanrefer} is a 3D visual grounding dataset with $51,583$ descriptions based on the $800$ \textit{ScanNet}~\cite{dai2017scannet} scenes.
Each scene has $13.81$ objects and $64.48$ descriptions on average. 
The evaluation metric of the dataset is the Acc@$m$IoU, which means the fraction of descriptions whose predicted box overlaps the ground truth with IoU $>m$, where $m\in\{0.25, 0.5\}$.
The accuracy is reported in \textit{unique} and \textit{multiple} categories.
Specifically, a target object is classified as \textit{unique} if it is the only object of its class in the scene; otherwise, it is classified as \textit{multiple}.

\vspace{0.5mm}
\noindent\textbf{Nr3D and Sr3D.}~The \textit{ReferIt3D} dataset~\cite{achlioptas2020referit3d} is also based on the \textit{ScanNet~}~\cite{dai2017scannet} scenes.
It contains two subsets: \textit{Sr3D} and \textit{Nr3D}.
\textit{Sr3D}~(Spatial Reference in 3D) contains $83,572$ synthetic expressions generated by templates and \textit{Nr3D}~(Natural Reference in 3D) consists of $41,503$ human expressions.
It directly provides segmented point clouds for each object as inputs rather than the whole scene.
The evaluation metric of \textit{ReferIt3D} is the accuracy, \ie, whether the model correctly selects the target among objects.

\subsection{Implementation Details}
Our model is trained end-to-end with the AdamW optimizer~\cite{DBLP:journals/corr/abs-1711-05101} and a batch size of $32$ for $32$ epochs.
The initial learning rates of TPM layers and the rest of the model are empirically set to $1e-4$ and $1e-3$, respectively.
We apply learning rate decay at epoch \{$16$, $24$, $28$\} with a rate of $0.1$. 
We adopt the pre-trained PointNet++~\cite{pointnet++} following the settings in ~\cite{groupfree} and the language encoder in~ \cite{radford2021learning}, while the rest of the network is trained from scratch.
For the \textit{ScanRefer} dataset, we use $xyz$ coordinates, RGB values, normal vectors, and extracted multiview features as inputs following~\cite{scanrefer}.
The number $M$ of $\mathbf{P}_{seed}$ is empirically set to $1024$.
The number $K_0$ of $\mathbf{P}_{0}$ is empirically set to $512$.
The number $T$ of TPM layers is set to $4$, and we select $50\%$ keypoints in each layer, \ie, $\left\{{K}_t|t \in \left\{1,2,3,4\right\}\right\} = \left\{256, 128, 64, 32\right\}$.
The loss weights are empirically set to $\alpha_1=0.1$, $\alpha_2=0.8$, $\alpha_3=5$, $\alpha_4=0.1$ for balancing terms.
We set $k_1$ to $4$, $\theta$ to $0.25$ in $\mathcal{L}_{\mathrm{VG}}$, and $k_2$ to $5$ in $\mathcal{L}_{\mathrm{DKS}}$.
All experiments are implemented with PyTorch on a single NVIDIA V100 GPU.

\subsection{Quantitative Comparison}

\begin{figure}[t]
    \centering 
    \includegraphics[width=0.43\textwidth]{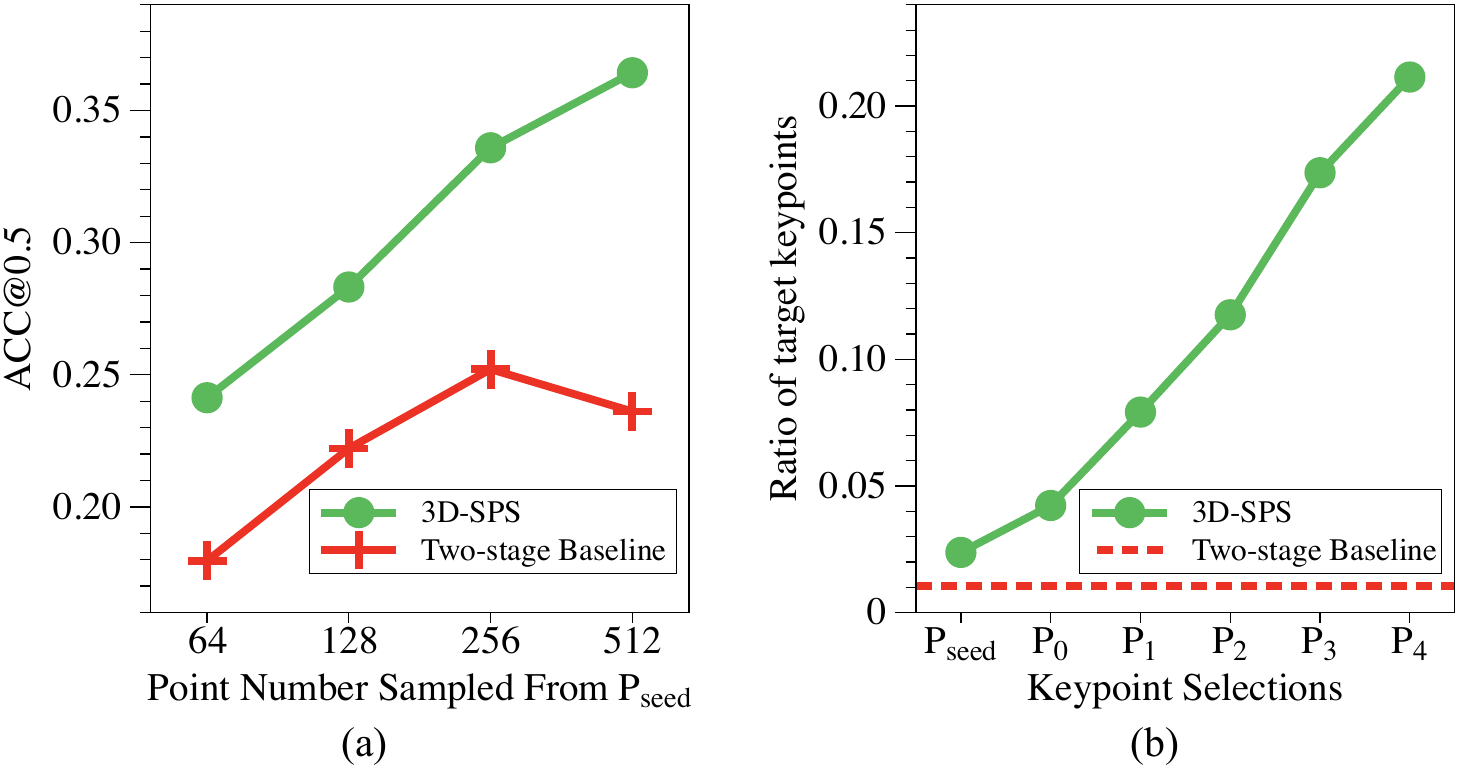}
    \caption{
    \noindent\textbf{Effectiveness Validation.}
    (a)~As the point number sampled from $\mathbf{P}_{seed}$ increases, our 3D-SPS performs better. The performance of the two-stage baseline first increases and then decreases.
    (b)~As the progressive language-relevant keypoint selection goes, the ratio of target keypoints in our 3D-SPS increases after each selection. Also, this ratio keeps outperforming language-irrelevant sampling~(\eg, FPS) used in the two-stage baseline.}
    \label{fig:exp-liner}
\end{figure}

In Table~\ref{tab:scanref-all-methods} and~\ref{tab:referit3d-all-methods}, we compare 3D-SPS with existing 3D VG works on \textit{ScanRefer} and \textit{Nr3D/Sr3D} datasets. 
The methods involved are 2D-based methods SCRC~\cite{hu2016natural} and One-stage~\cite{yang2019fast}, the segmentation-based two-stage methods TGNN~\cite{DBLP:conf/aaai/HuangLCL21} and InstanceRefer~\cite{Yuan_2021_ICCV}, the detection-based two-stage methods SAT~\cite{sat},
3DVG-Transformer~\cite{Zhao_2021_ICCV}, ScanRefer~\cite{scanrefer}, and ReferIt3DNet~\cite{achlioptas2020referit3d}. 
\vspace{0.5mm}
\noindent\textbf{ScanRefer.}~3D-SPS outperforms the existing methods by a large margin, as shown in Table~\ref{tab:scanref-all-methods}.
In the \textit{Input} column, \textit{3D only} stands for \textit{xyz + RGB + normals}, and \textit{2D + 3D} means an extra $128$-dimensional \textit{multiview} feature for each point is added to \textit{3D only}.
We concatenate these multiview features with our point features from the backbone and feed them into TPM together.
In the \textit{3D only} setting, 3D-SPS has improved by $+1.96\%$ at Acc$@0.5$ and $+1.75\%$ at Acc$@0.25$ compared to the existing state-of-the-art methods. 
In the \textit{2D+3D} setting, 3D-SPS outperforms the existing methods by $2.31\%$ at Acc$@0.5$ and $1.25\%$ at Acc$@0.25$.

Note that TGNN and InstanceRefer both rely on a pre-fixed 3D instance segmentation model. Thus InstanceRefer performs better on the Acc@0.5 score in the \textit{Unique} subset.

\vspace{0.5mm}
\noindent\textbf{Nr3D \& Sr3D.}~The task of the \textit{ReferIt3D} dataset~(\textit{Nr3D} \& \textit{Sr3D}) is to identify the target object among the given ground truth object bounding boxes.
We modify 3D-SPS accordingly, removing DKS and only verifying the effectiveness of TPM.
For fair comparisons, we adopt 2D semantic assisted training proposed by SAT~\cite{sat} in the training process and only use 3D inputs in the inference process.
Results in Table~\ref{tab:referit3d-all-methods} show progressive selection is effective for referring tasks.
3D-SPS significantly improves the grounding accuracy by $+2.3\%$ in \textit{Nr3D} and $+4.7\%$ in \textit{Sr3D}.
Although LanguageRefer performs better on the \textit{Easy} subset of the synthetic dataset \textit{Sr3D}, 3D-SPS outperforms it by a large margin on the more challenging \textit{Hard} subset.

\vspace{0.5mm}
\noindent\textbf{Effectiveness Validation.}~
Figure~\ref{fig:exp-liner} confirms that our main idea, \ie, progressive keypoint selection, can address the issues from the motivation in Sec.~\ref{sec:intro}. 
We analyze 3D-SPS and the two-stage method baseline~\cite{scanrefer} on the entire validation set of \textit{ScanRefer}.
As shown in Figure~\ref{fig:exp-liner}~(a), the two-stage baseline faces the dilemma of the point number sampled from $\mathbf{P}_{seed}$. In contrast, 3D-SPS benefits from more sampled points.
According to Figure \ref{fig:exp-liner}~(b), the two-stage baseline is limited by the small ratio of target keypoints due to the language-irrelevant keypoint sampling, while the ratio in 3D-SPS increases significantly after each selection.

\begin{table}[t]
    \centering
    \resizebox{0.30\textwidth}{!}{
    \begin{tabular}{c|c|c}
        \toprule
        & Acc@0.25 & Acc@0.5\\
        \hline
        FPS                & $43.83$ & $31.88$ \\
        DKS~(w/o $\mathbf{s}_d$)    & $46.15$ & $34.95$ \\
        DKS~(w/o $\mathbf{s}_o$)    & $46.06$ & $35.19$ \\
        {DKS}       & $\mathbf{47.65}$ & $\mathbf{36.43}$ \\
        \bottomrule
    \end{tabular}
    }
    \caption{Ablations on the sampling strategy of DKS.}
    \label{tab:exp-2}\vspace{-4mm}
\end{table}

\begin{table}[t]
    \centering
    \resizebox{0.42\textwidth}{!}{
    \begin{tabular}{c|c|c|c|c|c}
        \toprule
        $T$ & $1$ & $2$ & $3$ & $4$ & $5$\\
        \hline
        Acc@0.25 & $45.37$ & $45.99$ & $46.48$ & $\mathbf{47.65}$ & $47.02$ \\
        Acc@0.5 & $33.13$ & $33.97$ & $34.53$ & $\mathbf{36.43}$ & $36.07$\\
        \bottomrule
    \end{tabular}
    }
    \caption{Ablations on the layer number $T$ in TPM.}\vspace{-5mm}
    \label{tab:exp-3}
\end{table}

\begin{table}[t]
    \centering
    \resizebox{0.47\textwidth}{!}{
    \begin{tabular}{c|c|c|c|c|c|c}
        \toprule
        Keypoints & \multicolumn{5}{c|}{w/o selection} & {w/ selection}\\
        \cline{2-7}
        Num & $32$ & $64$ & $128$ & $256$ & $512$ & {$512\rightarrow 32$} \\
        \hline
        Acc@0.25 & $42.06$ & $44.77$ & $46.30$ & $46.38$ & $46.09$ & $\mathbf{47.65}$\\
        Acc@0.5 &  $31.89$ & $33.88$ &$34.99$ & $35.53$ & $34.98$ & $\mathbf{36.43}$\\
        \bottomrule
    \end{tabular}
    }
    \caption{Ablations of TPM on whether to select keypoints and different keypoint numbers. Our default setting is \textit{w/ selection}, where we progressively select keypoints from $512$ to $32$.}
    \label{tab:exp-4}\vspace{-4mm}
\end{table}

\begin{figure*}[t]
    \centering 
    \includegraphics[width=0.93\textwidth]{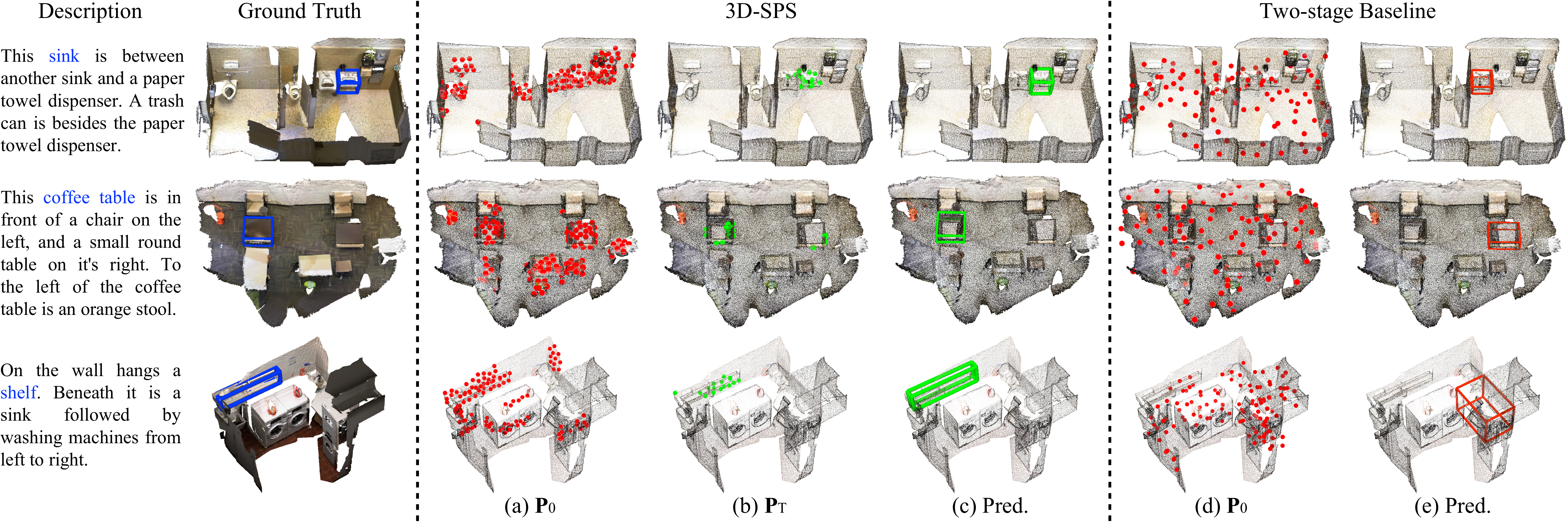}\vspace{-3mm}
    \caption{
The two-stage baseline~(ScanRefer) fails while our 3D-SPS predicts correctly since 3D-SPS can select more valuable keypoints. {(a)} Language-relevant keypoints $\mathbf{P}_0$ sampled by DKS. {(b)} Target keypoints $\mathbf{P}_T$ selected by TPM. {(c)} Bounding boxes predicted by 3D-SPS. {(d)} Language-irrelevant keypoints sampled by FPS. {(e)} Bounding boxes predicted by ScanRefer.}
    \label{fig:exper-1}
\end{figure*}

\begin{figure*}[t]
    \centering 
    \includegraphics[width=0.93\textwidth]{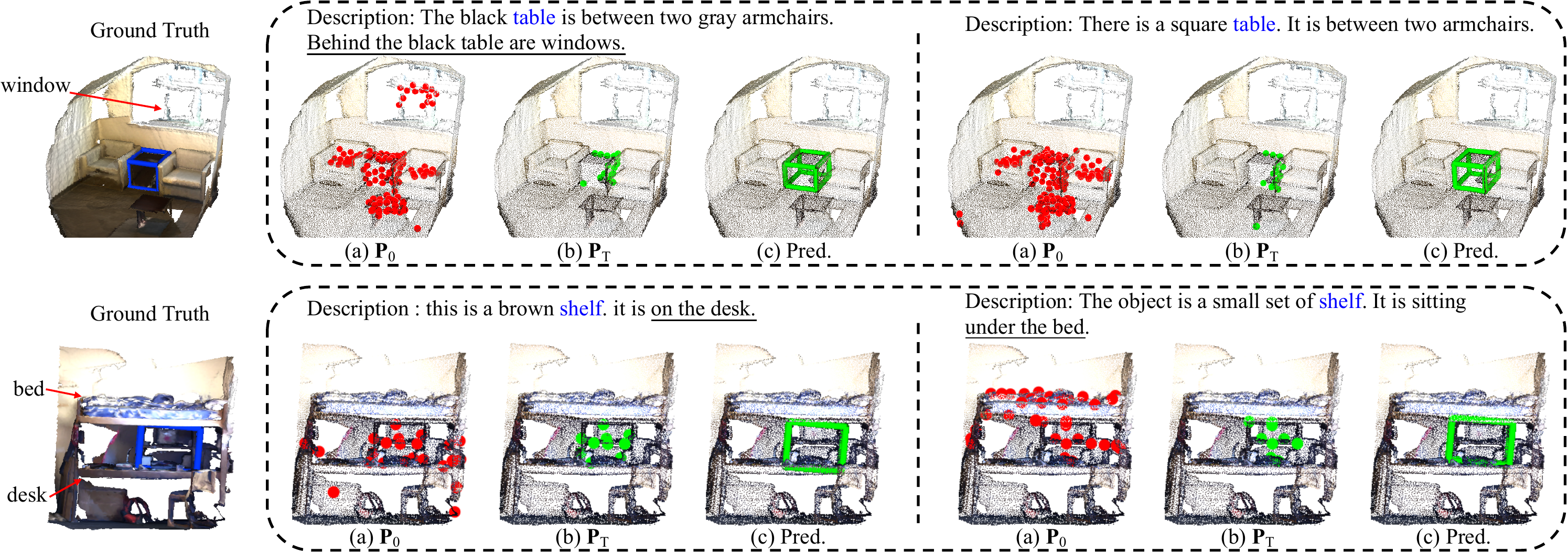}
    \caption{Visualization of the same referring target with different descriptions in 3D-SPS. (a) $\mathbf{P}_{0}$ sampled by DKS. Comparing the left and right subfigures in each row, when the language-relevant objects change~(\eg, \textit{window}, \textit{desk}, \textit{bed}), 3D-SPS focuses on different keypoints~(\textcolor{red}{red} keypoints). (b) $\mathbf{P}_{T}$ selected by TPM. {(c)} The predicted target bounding box.
    }\vspace{-3mm}
    \label{fig:exper-2}
\end{figure*}

\subsection{Ablation Study}
In this subsection, we investigate the contribution of the proposed DKS and TPM module.
We take \textit{ScanRefer} as an example and report the \textit{Overall} accuracy in \textit{3D only} setting.

\vspace{0.5mm}
\noindent\textbf{Sampling Strategy of DKS.}~
Table~\ref{tab:exp-2} shows the ablations of sampling strategy in the DKS module.
FPS~\cite{pointnet++} is a widely adopted point sampling method, which 
makes an effort to cover the whole scene without special attention to the language-relevant points.
\textit{DKS (w/o $\mathbf{s}_d$)} means that only the object confidence score $\mathbf{s}_o$ is utilized, and \textit{DKS (w/o $\mathbf{s}_o$)} represents that only the description relevance score $\mathbf{s}_d$ is used.
\textit{DKS} means that both $\mathbf{s}_o$ and $\mathbf{s}_d$ are adopted and is the full version of the proposed DKS module.
According to the results in Table~\ref{tab:exp-2}, $\mathbf{s}_o$ and $\mathbf{s}_d$ are both beneficial to the referring task, helping DKS select description-related keypoints near object centers. The joint use of $\mathbf{s}_o$ and $\mathbf{s}_d$ can produce promising results.

\vspace{0.5mm}
\noindent\textbf{Layer Number of TPM.}~
We investigate the performance on different TPM layer numbers $T \in \{1,2,3,4,5\}$. As shown in Table~\ref{tab:exp-3}, more TPM layers bring higher accuracy, which demonstrates that TPM and the progressive mining are essential to grounding. We take $T=4$ as the default setting since more layers might force the model to leave out some keypoints of the target object and miss the best bounding box.

\vspace{0.5mm}
\noindent\textbf{Progressive Selection of TPM.}~
To further confirm the effectiveness of progressive keypoint selection, we compare the results on whether to adopt keypoint selection, as shown in Table~\ref{tab:exp-4}. 
In detail, for the \textit{w/o selection} setting, we only conduct multi-modal self/cross-attention. 
In this way, the number of keypoints does not change in TPM, and the predicted box is chosen from all keypoints after TPM. 
From Table~\ref{tab:exp-4}, with the increase of keypoint numbers, the performance of the \textit{w/o selection} setting rises at first and then declines. 3D-SPS~(\textit{w/ selection}) achieves significant improvement compared to the \textit{w/o selection} settings. 
This observation proves the benefits of progressive keypoint selection.

\subsection{Qualitative comparison}
In this subsection, we perform a qualitative comparison on \textit{ScanRefer} validation set to show how 3D-SPS works.

\noindent\textbf{Language-relevant Keypoints.}~
We visualize the progressive keypoint selection process of 3D-SPS in Figure~\ref{fig:exper-1} and compare it with the two-stage baseline ScanRefer~\cite{scanrefer}.
Enabled by DKS and TPM, 3D-SPS gradually focuses on the target. In contrast, the attention of ScanRefer is scattered everywhere in the scene and ultimately fails to locate the target due to the separation of detection and matching. 

\noindent\textbf{Language-adapted Keypoints.}~
3D-SPS selects different keypoints for the same target with different descriptions. 
As shown in Figure~\ref{fig:exper-2}~(upper), to locate the \textit{table}, 3D-SPS selects some keypoints on the \textit{window} for subsequent mining when \textit{window} is mentioned in the left sample. On the right, when only \textit{armchairs} is mentioned, 3D-SPS only selects keypoints on \textit{armchairs} and \textit{tables}.
In Figure~\ref{fig:exper-2}~(lower), for the target \textit{shelf}, 3D-SPS finds more keypoints related to the \textit{desk} when the shelf is described as \textit{on the desk} in the left sample. When the description contains \textit{under the bed}, the model pays more attention to the \textit{bed}.

%% file: tex/5_conclusion.tex
\vspace{-2mm}
\section{Conclusion and Discussion}
\vspace{-1mm}
In this work, we propose a brand new 3D visual grounding framework on point clouds named 3D Single-Stage Referred Point Progressive Selection method~(3D-SPS).
Under the guidance of language, it progressively selects keypoints following a coarse-to-fine pattern and directly localizes the target at a single stage.
Comprehensive experiments reveal that our method outperforms the existing 3D VG methods on both \textit{ScanRefer} and \textit{Nr3D / Sr3D} datasets by a large margin, leading to the new state-of-the-art performance.

\noindent\textbf{Limitation.} The limitation of 3D-SPS exists due to the complexity of 3D point clouds and free-form description, although we have made significant improvements on existing methods. The view-dependent descriptions and the ambiguous queries can both confuse the model. 
These limitations could guide our future work.

\vspace{2mm}
\noindent\textbf{Acknowledgement.}~This research is partly supported by National Natural Science Foundation of China (Grant 62122010, 61876177), Fundamental Research Funds for the Central Universities, and Key R \& D Program of Zhejiang Province(2022C01082).

\newpage